\documentclass[3p,times,procedia]{elsarticle}
\usepackage{ecrc}
\usepackage{amsmath}

\volume{00}

\firstpage{1}

\journalname{Transportation Research Procedia}

\runauth{Milan Straka}

\jid{trpro}

\usepackage{amssymb}
\biboptions{authoryear}

\usepackage[figuresright]{rotating}
\usepackage{float} 
\usepackage{tabularx}
\usepackage{xcolor}
\usepackage{bm}

\usepackage{caption}
\usepackage{subcaption}

\usepackage[bookmarks=false]{hyperref}
    \hypersetup{colorlinks,
      linkcolor=blue,
      citecolor=blue,
      urlcolor=blue}

\begin{document}
\begin{frontmatter}

%% Title, authors and addresses

%% use the tnoteref command within \title for footnotes;
%% use the tnotetext command for the associated footnote;
%% use the fnref command within \author or \address for footnotes;
%% use the fntext command for the associated footnote;
%% use the corref command within \author for corresponding author footnotes;
%% use the cortext command for the associated footnote;
%% use the ead command for the email address,
%% and the form \ead[url] for the home page:
%%
%% \title{Title\tnoteref{label1}}
%% \tnotetext[label1]{}
%% \author{Name\corref{cor1}\fnref{label2}}
%% \ead{email address}
%% \ead[url]{home page}
%% \fntext[label2]{}
%% \cortext[cor1]{}
%% \address{Address\fnref{label3}}
%% \fntext[label3]{}

\dochead{14th International scientific conference on sustainable, modern and safe transport}%

\title{A matrix approach to detect temporal behavioral patterns at electric vehicle charging stations}

%% use optional labels to link authors explicitly to addresses:
%% \author[label1,label2]{<author name>}
%% \address[label1]{<address>}
%% \address[label2]{<address>}

\author[a]{Milan Straka} 
\author[a]{Lucia Piatrikov\'{a}}
\author[b]{Peter van Bokhoven}
\author[a]{\v{L}ubo\v{s} Buzna}
\address[a]{University of \v{Z}ilina, Univerzitn\'{a} 8215/1, \v{Z}ilina, Slovakia}
\address[b]{ElaadNL, Utrechtseweg 310 (bld. 42B), 6812 AR Arnhem (GL), The Netherlands}
\begin{abstract}
%% Text of abstract
Based on the electric vehicle (EV) arrival times and the duration of EV connection to the charging station, we identify charging patterns and derive groups of charging stations with similar charging patterns applying two approaches. The ruled based approach derives the charging patterns by specifying a set of time intervals and a threshold value. In the second approach, we combine the modified l-p norm (as a matrix dissimilarity measure) with hierarchical clustering and apply them to automatically identify charging patterns and groups of charging stations associated with such patterns. A dataset collected in a large network of public charging stations is used to test both approaches. Using both methods, we derived charging patterns. The first, rule-based approach, performed well at deriving predefined patterns and the latter, hierarchical clustering, showed the capability of delivering unexpected charging patterns.
\end{abstract}

\begin{keyword}
electromobility, clustering, charging stations, data analysis; 

%% keywords here, in the form: keyword \sep keyword

%% PACS codes here, in the form: \PACS code \sep code

%% MSC codes here, in the form: \MSC code \sep code
%% or \MSC[2008] code \sep code (2000 is the default)

\end{keyword}
\cortext[cor1]{Corresponding author. Tel.: +421-41-553-4210.}
\end{frontmatter}

%\correspondingauthor[*]{Corresponding author. Tel.: +0-000-000-0000 ; fax: +0-000-000-0000.}
\email{milan.straka@fri.uniza.sk}

\setlength{\belowdisplayskip}{6pt} \setlength{\belowdisplayshortskip}{6pt}
\setlength{\abovedisplayskip}{6pt} \setlength{\abovedisplayshortskip}{6pt}

\section{Introduction}
 
In the Netherlands, there are currently operated thousands of public charging stations, and the number of electric vehicles is steadily growing. Charging of EVs should be affordable and have no adverse effects on local power grids (e.g., increased peak demand)~\cite{van2019mobility}. Therefore, it is crucial to monitor and analyze utilization of charging stations to adapt them to actual conditions and optimize the interrelated systems. Thus, an adequate analysis of charging patterns is required. 

\subsection{Motivation}

Smart charging is an intelligent approach in which electric vehicles and charging stations share data, and use them to operate the stations more effectively. Some of the goals are to reduce peak demand and improve system stability. It is essential for smart charging to predict charging station users' behavior to maintain the power grid system's stability. When reliable predictions of how long the EVs are connected to the charging station are possible, we can reasonably distribute charging over the whole connection duration~\cite{van2019mobility, miyazaki2020clustering}.

Locations of public charging stations influence their usage. For example, charging stations in industrial areas are typically used during the working time, while charging stations in residential areas are used overnight. Thus, the charging stations may significantly differ in arrival times of EVs and connections duration. The application of prediction models to objects with similar behavior can improve their accuracy. Hence, the combination of clustering (to identify similar charging stations) with supervised learning can lead to desired outcomes and possibly improve smart charging technologies~\cite{miyazaki2020clustering}.

Here, we use the EV arrival times together with the duration of connections to represent the charging patterns at charging stations. Then, we cluster charging stations based on charging patterns and interpret the identified clusters. Refs. ~\cite{miyazaki2020clustering,develder2016quantifying, helmus2020data} also used a similar approach in other contexts.

\subsection{Literature review}
Recent Ref.~\cite{shahriar2020machine} provides a comprehensive review of supervised and unsupervised machine learning and deep learning methods for EV charging behavior analysis and prediction. Among other things, this article suggests as a future research direction the comprehensive cluster analysis of EV charging behavior and the use of reinforcement learning for EV scheduling. In the  paper~\cite{straka2019clustering}, clustering algorithms were applied to the charging stations while testing two approaches: Aggregation first - Clustering second and Clustering first - Categorization second. Four groups of charging stations were identified, where the temporal patterns play the leading role among the used indicators.  In~\cite{miyazaki2020clustering}, authors analyzed the EV's behavior and conducted a one-day ahead prediction with a pre-processing while utilizing hierarchical clustering with Euclidean distance. They predicted when and for how long the EVs would be connected the next day. Authors also considered the day of the week in the analysis. The accuracy of the prediction was improved by clustering vehicles and building a prediction model for each cluster compared with one prediction model built for all vehicles. In~\cite{helmus2020data} authors used a data-driven two-step clustering approach, firstly clustering the charging sessions and after that portfolios of charging sessions. They discovered and described various user types. The authors used four attributes, namely EV plugin time, connection duration, the time between consecutive charging sessions, and spatial distance between charging session locations of EV users to cluster the charging transactions. 

\subsection{Our contribution}
This paper's primary goal is to demonstrate the ability of clustering methods to identify temporal charging patterns at EV charging stations and to explore similar groups of charging stations. We represent the data using a charging matrix. First, we apply simple rules to the charging matrix to identify groups of charging stations that follow the pre-defined behavioral charging patterns. Second, we explore parameter values of a matrix similarity measure that could be used to automatically identify the charging patterns in combination with standard clustering methods. At the same time, similar charging stations are merged into the same cluster. This is the first paper applying a charging matrix and clustering methods to analyze the temporal patterns at charging stations to the best of our knowledge.

\section{Data}

\subsection{Dataset}
In this study, a public charging infrastructure dataset provided by the Dutch innovation company EVnetNL was used. It contains more than $1700$ charging stations located across the Netherlands and over one million charging transactions records. The data describes EV arrivals, departures, connection duration, etc. As the number of charging stations was not stable across the whole period covered by the dataset, we used only charging transactions that occurred in the year 2015.

\subsection{Data pre-processing}
We merged data from charging stations that were located close to each other (closer than 30 meters) into a single charging station, as these cases typically represent a situation when multiple stations form a charging pool. Nevertheless, in what follows, we will be further using the term charging station. To assure that a sufficient number of transactions represents all charging stations, we omitted from the analysis charging stations with less than $30$ transactions. After these steps, $1266$ charging stations remained, with $288$ charging transactions per charging station on average.
To get a more compact representation of data and consider that most of the transactions are shorter than $24$ hours, we omit all the transactions with connection time longer than $24$ hours. Hence, to represent charging sessions, we create a charging matrix $A \in \mathbb{R}^{24 \times 24}$ for each charging station. The rows represent the duration of a connection in hours, and columns represent the hour of the day of the corresponding EV arrival. In the cell $A_{i,j}$ we store the empirical probability of observing a transaction starting in-between $i-1$-th and $i$-th hour of the day and having the duration from the interval $\langle j-1, j)$, measured in hours. All transactions with the connection time equal or longer than $24$ hours have been discarded.
 
\begin{figure}[H]
    \centering
    \includegraphics[width=0.5\textwidth]{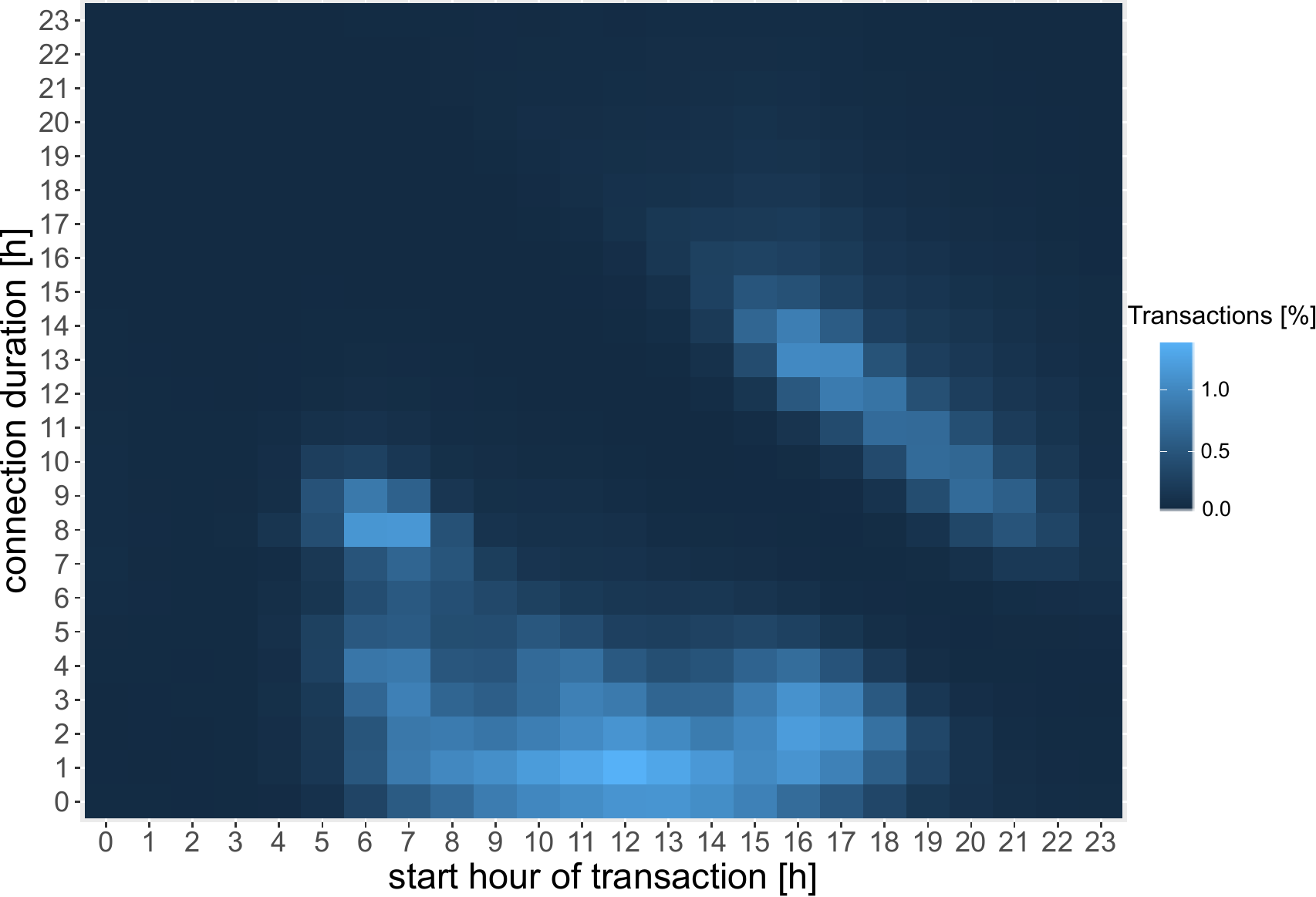}
    \caption{Heatmap of the charging matrix that was created from transactions of all considered charging stations. }
    \label{fig:all_charging_matrix}
\end{figure}

To provide a brief overview of the data, in Tables~\ref{tab:element_freqs} we present the frequencies of values in all charging matrices and in Figure~\ref{fig:all_charging_matrix} we show the heatmap obtained from the charging matrix that was created by considering the data from all charging stations. Many matrix cells that contain zero values and non-zero values are concentrated in clusters forming a complex charging pattern. 
\begin{table}[H]
\centering
\caption{Frequencies of values in charging matrices.}
\begin{tabular}{lllll}
\hline
& $A_{ij} = 0$ & $A_{ij} \in (0, 0.01\rangle $ & $A_{ij} \in (0.01, 0.1\rangle $ & $A_{ij} \geq 0.1 $ \\
\hline
\hline
frequency & 608496 & 79997 & 40506 & 237 \\
\hline
\end{tabular}
\label{tab:element_freqs}
\end{table}

\section{Methods}

First, we apply rule-based clustering, where the charging stations are clustered based on pre-set rules. We use a matrix dissimilarity measure to automate the clustering process and possibly identify unexpected types of charging patterns combined with hierarchical clustering.

\subsection{Rule-based clustering} 
This method utilizes specified time intervals that are applied to the start time and connection time (i.e., rows and columns of the charging matrix) and defines submatrices of the charging matrices. The sum of elements in a submatrix is then compared with a threshold value of $\theta$. A charging pattern is given by the layout of submatrices with the sum of values higher than $\theta$. Finally, charging stations are grouped by their charging patterns. 

\subsection{Matrix dissimilarity measure}

\begin{figure}[H]
    \centering
    \includegraphics[width=0.4\textwidth]{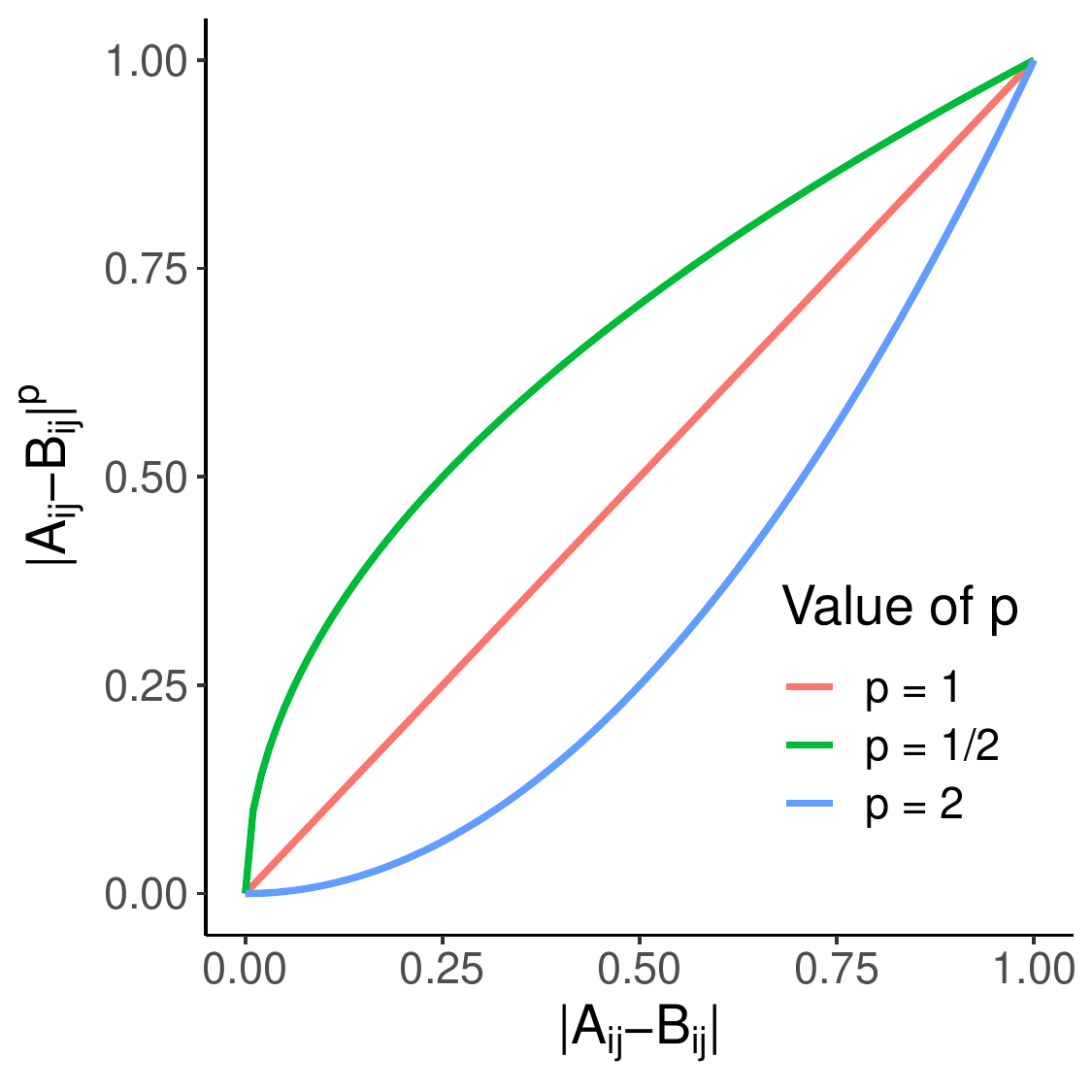}
    \caption{Comparison of values $|A_{ij}-B_{ij}|^p$ as a function of $|A_{ij}-B_{ij}|$  for $p = \{\frac{1}{2}, 1, 2\}$.}
    \label{fig:distanceCompare}
\end{figure}

To cluster charging matrices, a measure that could quantify dissimilarity of matrices is required. As the elements of the matrices at the same position represent the same temporal occurrence, we quantify the matrix similarity as the sum of pairwise terms, one for each element of the charging matrix.
Slightly modifying the $l_{p}$ norm (where $o = p$) and applying it to the matrix format, we define the following measure 
\begin{equation}
     d(A,B,o,p) = \sqrt[o]{\sum_{i=1}^{24}\sum_{j=1}^{24}|A_{ij}-B_{ij}|^p},
     \label{eq:L_norm}
\end{equation}
where $A$ and $B$ are charging matrices and $p \geq 0$ and $o \geq 1$ are parameters. As the elements of a charging matrix are probability values that sum to one, the expression $|A_{ij}-B_{ij}|$ must be between zero and one (including zero and one). The frequencies of values in Table~\ref{tab:element_freqs} suggest that we can expect the expression $|A_{ij}-B_{ij}|$ to take in most of the cases value zero or a value that is very close to zero. For these reason it appears as more suitable choice $p \leq 1$, which makes the dissimilarity measure more sensitive to values close to zero. To provide an intuition on this issue, in Figure~\ref{fig:distanceCompare}, we compare values of $|A_{ij}-B_{ij}|^p$ as a function of $|A_{ij}-B_{ij}|$ for $p\in \{\frac{1}{2}, 1, 2\}$. For $p = 1/2$ the value $|A_{ij}-B_{ij}|^p$ grows faster, making it more sensitive to small values of $|A_{ij}-B_{ij}|$.

\subsection{Clustering methods}

Clustering is a process that assigns objects into disjoint subgroups, so in one group, we find objects more similar to each other than to objects belonging to the other groups. 

The agglomerative hierarchical clustering assigns first all objects to individual clusters, and in the next steps, pairs of clusters are iteratively merged until all observations belong into a single cluster. Such a process leads to a hierarchy of clusters, which is given by the order in which the clusters were merged. A dendrogram is a natural tree-based representation of this hierarchy~\cite{james2013introduction}.

We use hierarchical clustering as it does not require to decide about the number of clusters in advance. Instead, we can analyze the nesting of clusters in the dendrogram and observe the composition of clusters of various numbers. Additionally, the nesting provides an additional intuition about the relationship between stations and clusters. 

Clusters might be analyzed by choosing a representation of the cluster. We consider two approaches. As the first approach, we create a matrix representing a cluster by a normalized element-wise sum of all charging matrices belonging to the cluster. The second approach selects as the representative of a cluster a median charging matrix with the smallest sum of element-wise distances to all other charging matrices in the cluster.

\section{Results}

\subsection{Rule-based clustering}

As the first step, we explored the charging matrices by eyeballing the heatmaps and discussions with experts from ElaadNL we identified some regular charging patterns similar to those described in~\cite{develder2016quantifying}\, e.g., work charging where EVs are connected to charging stations in the morning and disconnected in the evening, home charging where EVs are connected in the evening and remain until night or morning next day.

\begin{figure}[H]
    \centering
    \includegraphics[width=0.4\textwidth]{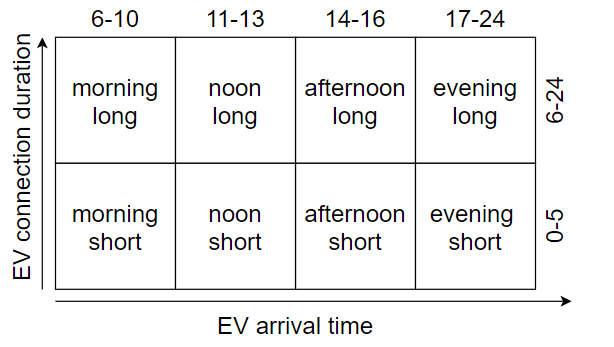}
    \caption{Partitioning of the charging matrix according to the EV arrival time and connection time.}
    \label{fig:MeasuresScheme}
\end{figure}

These patterns can be captured by dividing the studied interval for the EV arrival times into four parts corresponding to the morning, from 4 to 10, noon from 11 to 13, afternoon from 14 to 16, and evening from 17 to 23. Furthermore, to express observed patterns, we divide the connection time into two intervals corresponding to short (less than 6 hours) and long duration (more than 6 hours). Submatrices, which are obtained by such partitioning, are displayed in Figure~\ref{fig:MeasuresScheme}.

\begin{figure}[H]
    \centering
    \includegraphics[width=0.97\textwidth]{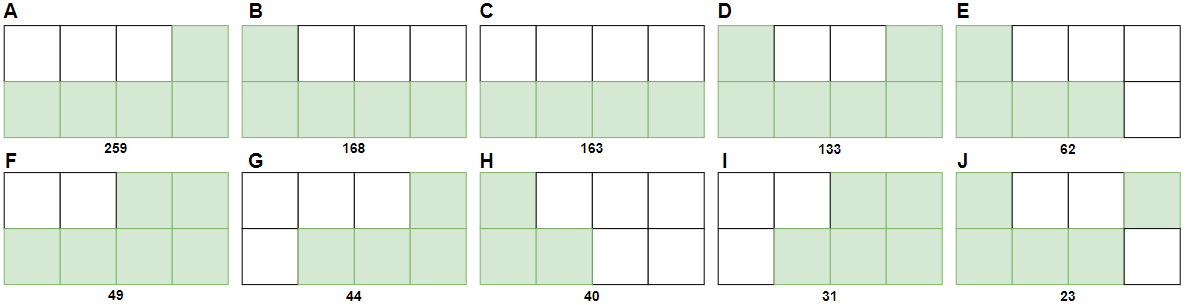}
    \caption{Resulting patterns obtained by the rule-based categorization (the presence of the pattern is indicated with green filled squares). The number of charging matrices belonging to each cluster is displayed below each pattern.}
    \label{fig:MeasuresCompare}
\end{figure}

The threshold $\theta$ that decides about the affiliation of a charging matrix to a charging pattern we set to a value of $0.06$. If the sum of probabilities in a submatrix exceeds $\theta$, a charging pattern corresponding to a given submatrix is considered to be present. We explored the interval $\langle 0.03, 0.15 \rangle$ and $\theta = 0.06$ returned the results that we considered as the most meaningful. The charging matrices exhibiting the same charging pattern were assigned to the same cluster. 
We considered a split of the charging into $8$ submatrices, thus the maximum number of clusters is $256$. We obtained $75$ clusters and $10$ largest clusters, together with their frequencies are presented in Figure~\ref{fig:MeasuresCompare},  Altogether, top-ten clusters contain around $75\%$ of all the charging matrices.

Patterns A-G, I, and J presented in Figure~\ref{fig:MeasuresCompare} contain predominantly short charging, which is occasionally combined with morning or afternoon long charging. This pattern suggests that they could correspond to various types of home charging.  In the case of pattern H, the charging activity is concentrated in the morning, only what corresponds to work charging.

\subsection{Hierarchical clustering}

In the experiments presented in this section, we apply the previously described matrix dissimilarity measure combined with agglomerative hierarchical clustering, utilizing complete-linkage. We explored the following combination of parameter values: $p = o;$ while $o$ and $p \in \{1, 2, 3\}$ and $o = 1$ while $p \in \{\frac{1}{2}, \frac{1}{3}, \frac{2}{3}, 2, 3 \}$. We limited the exploration of the dendrograms to the number of clusters ranging from $2$ to $10$ .

In Table~\ref{tab:clust_exp}, we present only the size of clusters that were obtained at the height $h$ of the dendrogram that splits the observations into 10 clusters. The frequencies in clusters are arranged in decreasing order. 

\begin{table}[H]
\centering
\caption{The clusters' size obtained when we chose 10 clusters for each selected combination of parameters $o$ and $p$.}
\begin{tabular}{lllllllllll}
\hline
    \multicolumn{11}{c}{Cluster} \\
                     & 1    & 2   & 3   & 4   & 5   & 6  & 7  & 8  & 9 & 10 \\ \hline
                     \hline
$\bm{o = 1, p = \frac{2}{3}}$   & \textbf{420}  & \textbf{209} & \textbf{193} & \textbf{172} & \textbf{160} & \textbf{46} & \textbf{42} & \textbf{19} & \textbf{3} & \textbf{2}  \\
$o = 1, p = \frac{1}{2}$   & 522  & 265 & 264 & 132 & 45  & 29 & 5  & 2  & 1 & 1  \\
$o = 1, p = \frac{1}{3}$   & 613  & 211 & 198 & 133 & 95  & 8  & 3  & 3  & 1 & 1  \\
$o = 1, p = 2$               & 1166 & 69  & 14  & 8   & 3   & 2  & 1  & 1  & 1 & 1  \\
$o = 1, p = 3$               & 1253 & 3   & 2   & 2   & 1   & 1  & 1  & 1  & 1 & 1  \\
$\bm{o = p = 1}$                & \textbf{268}  & \textbf{254} & \textbf{241} & \textbf{197} & \textbf{167} & \textbf{95} & \textbf{24} & \textbf{13} & \textbf{6} & \textbf{1}  \\
$o = p = 2$                & 1166 & 69  & 14  & 8   & 3   & 2  & 1  & 1  & 1 & 1  \\
$o = p = 3$                & 1253 & 3   & 2   & 2   & 1   & 1  & 1  & 1  & 1 & 1  \\ \hline
\end{tabular}
\label{tab:clust_exp}
\end{table}

\begin{figure}[H]
    \centering
    \includegraphics[width=0.97\textwidth]{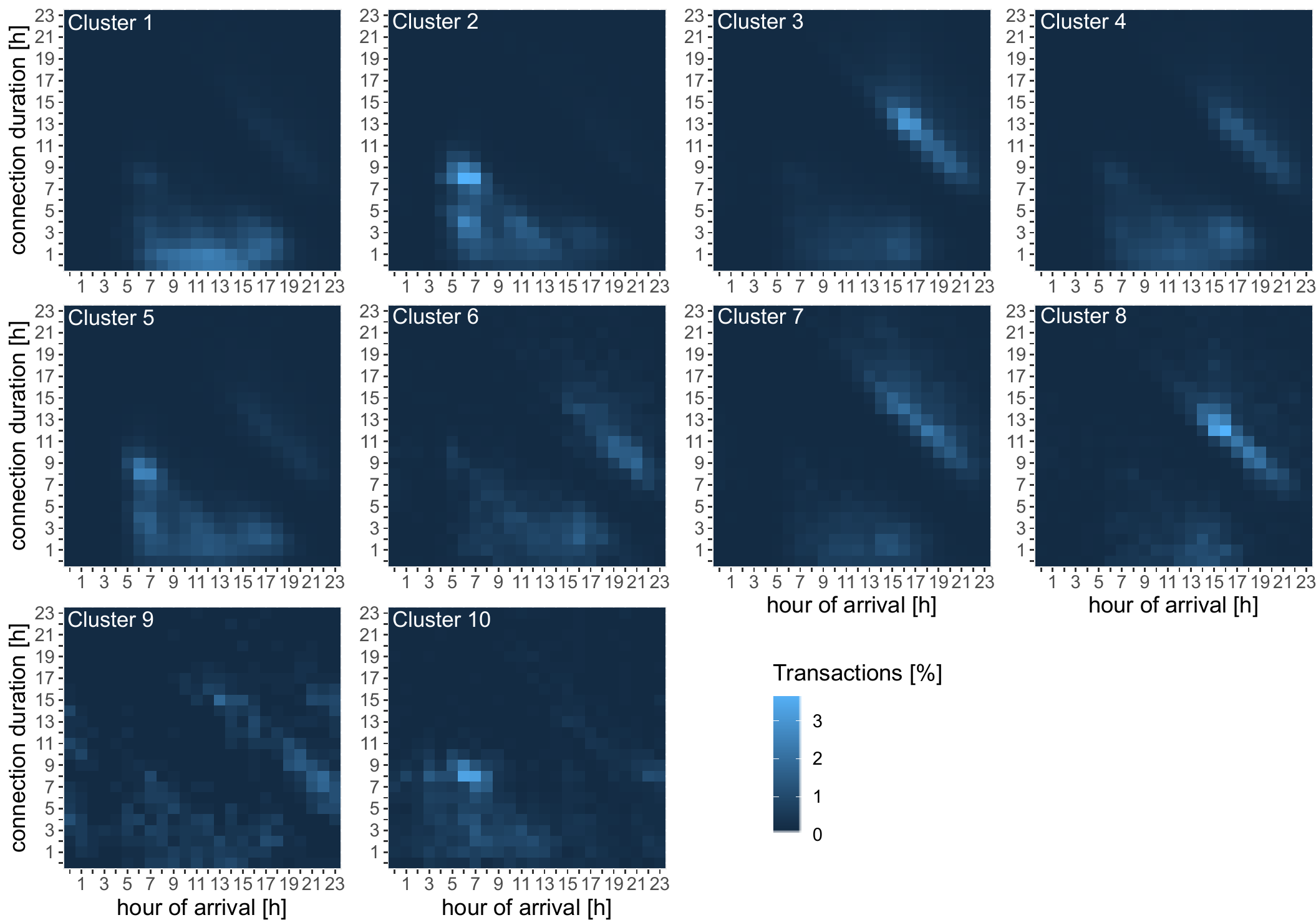}
    \caption{Heatmaps visualizing the normalized element-wise sum of all charging matrices belonging each of 10 clusters identified by hierarchical clustering.}
    \label{fig:clust_3_results}
\end{figure}

We obtain a favorable clustering result when all clusters are populated evenly. This requirement is best satisfied by the dissimilarity measure  with parameter values $o = 1, p = \frac{2}{3}$ and $o = p = 1$ whose corresponding rows we highlighted in bold. By eyeballing each cluster's median charging matrix and analyzing the distribution of observations among the clusters, we have chosen $o = 1, p = \frac{2}{3}$ as giving the most favorable outcome. For this case, in Figure~\ref{fig:clust_3_results}, we present the normalized element-wise sum of all charging matrices belonging to the 10 clusters. The first cluster contains mostly short charging, the second prevalently morning long charging, and is similar to the fifth and tenth cluster, where all these four clusters have a low amount of evening charging. The third and eighth clusters contain mostly long evening charging, which could be suitable for smart charging purposes. The rest of the clusters combine all types of charging in approximately uniform proportions.

\section{Discussion and conclusions}
As expected, the rule-based categorization can cluster similar charging stations based on pre-defined charging patterns very well but lacks the ability to identify unexpected charging patterns. Therefore, we complemented the rule-based clustering with hierarchical clustering. By numerical experiments, we demonstrated that both these methods are able to identify common charging patterns. 

Our results might have various applications, e.g., they could be used to enhance the prediction models of the temporal behavior at the charging infrastructure. The proposed methodology could also be used to identify temporal behavioral charging patterns of EV drivers, who could be in the analyses represented in a similar way as charging stations.

\section*{Acknowledgement}
This research was supported in part by the research projects VEGA 1/0089/19 “Data analysis methods and decisions support tools for service systems supporting electric vehicles,” APVV-19–0441 “Allocation of limited resources to public service systems with conflicting quality criteria”, by the Slovak Research and Development Agency under the contract no. SK-IL-RD-18–005 Operational Program Integrated Infrastructure 2014–2020 “Innovative solutions for propulsion, power and safety components of transport vehicles” code ITMS313011V334, co-financed by the European Regional Development Fund.
\bibliographystyle{abbrv}

\begin{thebibliography}{1}
	
	\bibitem{develder2016quantifying}
	C.~Develder, N.~Sadeghianpourhamami, M.~Strobbe, and N.~Refa.
	\newblock Quantifying flexibility in ev charging as dr potential: Analysis of
	two real-world data sets.
	\newblock In {\em 2016 IEEE International Conference on Smart Grid
		Communications (SmartGridComm)}, pages 600--605. IEEE, 2016.
	
	\bibitem{helmus2020data}
	J.~R. Helmus, M.~H. Lees, and R.~van~den Hoed.
	\newblock A data driven typology of electric vehicle user types and charging
	sessions.
	\newblock {\em Transportation Research Part C: Emerging Technologies},
	115:102637, 2020.
	
	\bibitem{james2013introduction}
	G.~James, D.~Witten, T.~Hastie, and R.~Tibshirani.
	\newblock {\em An introduction to statistical learning}, volume 112.
	\newblock Springer, 2013.
	
	\bibitem{miyazaki2020clustering}
	K.~Miyazaki, T.~Uchiba, and K.~Tanaka.
	\newblock Clustering to predict electric vehicle behaviors using state of
	charge data.
	\newblock In {\em 2020 IEEE International Conference on Environment and
		Electrical Engineering and 2020 IEEE Industrial and Commercial Power Systems
		Europe (EEEIC/I\&CPS Europe)}, pages 1--6. IEEE, 2020.
	
	\bibitem{shahriar2020machine}
	S.~Shahriar, A.~Al-Ali, A.~H. Osman, S.~Dhou, and M.~Nijim.
	\newblock Machine learning approaches for ev charging behavior: A review.
	\newblock {\em IEEE Access}, 8:168980--168993, 2020.
	
	\bibitem{straka2019clustering}
	M.~Straka and L.~Buzna.
	\newblock Clustering algorithms applied to usage related segments of electric
	vehicle charging stations.
	\newblock {\em Transportation Research Procedia}, 40:1576--1582, 2019.
	
	\bibitem{van2019mobility}
	R.~van~den Hoed, S.~Maase, J.~Helmus, R.~Wolbertus, Y.~el~Bouhassani, J.~Dam,
	M.~Tamis, and B.~Jablonska.
	\newblock E-mobility: getting smart with data.
	\newblock 2019.
	
\end{thebibliography}

\end{document}